\documentclass[sn-mathphys,Numbered]{sn-jnl}%
\usepackage{graphicx}%
\usepackage{multirow}%
\usepackage{amsmath,amssymb,amsfonts}%
\usepackage{amsthm}%
\usepackage{mathrsfs}%
\usepackage[title]{appendix}%
\usepackage{xcolor}%
\usepackage{textcomp}%
\usepackage{manyfoot}%
\usepackage{booktabs}%
\usepackage{algorithm}%
\usepackage{algorithmicx}%
\usepackage{algpseudocode}%
\usepackage{qcircuit}
\usepackage{braket}
\usepackage{float}
\usepackage{subcaption}
\usepackage{mathtools}
\usepackage{dirtytalk}

\usepackage{listings}%

\usepackage[breakable]{tcolorbox}
\tcbuselibrary{skins}
\usepackage{algpseudocode}

\theoremstyle{thmstyleone}%

\theoremstyle{thmstyletwo}%

\theoremstyle{thmstylethree}%


\begin{document}

\title[Article Title]{Federated Quantum Long Short-term Memory (FedQLSTM)}

\author*[1]{\fnm{Mahdi} \sur{Chehimi}}\email{mahdic@vt.edu}

\author[2]{\fnm{Samuel} \sur{Yen-Chi Chen}}

\author[1]{\fnm{Walid} \sur{Saad}}

\author[2]{\fnm{and Shinjae} \sur{Yoo}}

\affil[1]{\orgdiv{Wireless@VT, Bradley Department of Electrical and Computer Engineering}, \orgname{Virginia Tech}, \orgaddress{\city{Arlington}, \state{Virginia}, \postcode{22203}, \country{USA}}}

\affil[2]{\orgdiv{Computational Science Initiative}, \orgname{Brookhaven National Laboratory}, \orgaddress{ \city{Upton}, \state{NY}, \postcode{11973}, \country{USA}}}


\abstract{Quantum federated learning (QFL) can facilitate collaborative learning across multiple clients using quantum machine learning (QML) models, while preserving data privacy. Although recent advances in QFL span different tasks like classification while leveraging several data types, no prior work has focused on developing a QFL framework that utilizes temporal data to approximate functions useful to analyze the performance of distributed quantum sensing networks. In this paper, a novel QFL framework that is the first to integrate quantum long short-term memory (QLSTM) models with temporal data is proposed. The proposed \emph{federated QLSTM (FedQLSTM)} framework is exploited for performing the task of function approximation. In this regard, three key use cases are presented: Bessel function approximation, sinusoidal delayed quantum feedback control function approximation, and Struve function approximation. Simulation results confirm that, for all considered use cases, the proposed FedQLSTM framework achieves a faster convergence rate under one local training epoch, minimizing the overall computations, and saving 25-33\% of the number of communication rounds needed until convergence compared to an FL framework with classical LSTM models.}


\maketitle
\section{\label{sec:Indroduction}Introduction}
The field of \emph{quantum machine learning (QML)} is witnessing an increased interest by researchers from academia, governments, and industry \cite{debenedictis2018future}. QML harnesses the intrinsic parallelism of quantum computers, offering enhanced processing and analysis of intricate high-dimensional data. This potentially allows for notable improvements in computational complexity and efficiency compared to classical machine learning (ML) approaches. In general, QML models are \emph{hybrid quantum-classical} models with an underlying structure of quantum gates and circuits that emulates the operational principles of classical ML models \cite{biamonte2017quantum}. A successful implementation of this concept is seen in variational quantum circuits (VQCs), where quantum gates within the QML models are controlled by classically-optimized parameters \cite{benedetti2019parameterized}. VQCs have demonstrated success in a myriad of applications, such as classification \cite{mitarai2018quantum,chen2020hybrid,schuld2018circuit}, sequence modeling \cite{chen2022quantum,bausch2020recurrent}, generative modeling \cite{stein2020qugan}, and reinforcement learning \cite{chen2022variational}. Key examples of VQC architectures that can outperform their classical counterparts include quantum convolutional neural networks (QCNNs) for classification tasks \cite{cong2019quantum,chen2020qcnn}, quantum recurrent neural networks (QRNNs) \cite{bausch2020recurrent}, and quantum long short-term memory (QLSTM) models \cite{chen2022quantum,chen2022reservoir} for temporal data.  

In particular, QLSTM models have the potential to outperform traditional long short-term memory (LSTM) models while reducing the number of training parameters and training complexity \cite{chen2022quantum}. QLSTM models achieve high efficacy in the analysis of temporal sequences and time-series datasets, a category that predominantly encompasses sensor-generated data, as well as information integral to natural language processing applications \cite{di2022dawn}. 

In this regard, quantum sensing is an emerging quantum technology that has great capabilities. In particular, quantum sensors include probes that can sense or measure certain phenomena, e.g., a stimulus electric or magnetic field, with a precision that is much higher than the corresponding classical sensors \cite{giovannetti2004quantum}. Thus, such sensors can provide temporal data that show their quantum state and how it evolves over time under the effect of a stimulus. Henceforth, quantum sensors can represent a fundamental source of temporal data that can be fed into ML/QML models so as to perform efficient learning of the stimulus characteristics. Naturally, QLSTM models can have a direct application in analyzing and processing both classical and quantum data stemming from quantum sensors. 

However, such quantum sensors are typically deployed across distributed quantum sensor networks (QSNs), where multiple quantum sensors jointly participate in collaborative tasks like atomic clock synchronization and positioning \cite{giovannetti2001quantum}. Such QSNs can be both classical and quantum networks, and they enable several learning tasks to be performed on the data of each quantum sensor \cite{chehimi2023roadmap,chehimi2021entanglement_rate_optimization}. Moreover, in QSNs, each sensor can benefit from the data available to other sensors to enhance its learning capabilities and task prediction accuracy. However, in many applications, particularly ones related to positioning, quantum sensor data contains sensitive information that must be kept private and securely shared. In this regard, federated learning (FL) can be leveraged to enhance the privacy of distributed QSNs, while enabling collaborative learning between distributed quantum sensors \cite{kairouz2021advances}. In particular, clients participating in an FL framework only share the parameters of their local training models with a centralized server, without sharing their private data. This enables collaborative learning between several clients as the server aggregates and updates the model parameters, while preserving the clients' privacy. Leveraging recent advances in VQCs, and given their suitability to quantum sensors, \emph{quantum federated learning (QFL)}, where clients have local VQC models, has direct applications in analyzing and enhancing the performance of distributed QSNs \cite{chehimi2023foundations}. However, several fundamental questions remain to be answered in this field: \emph{How do we incorporate time series-based VQCs, like QLSTM, in QFL frameworks to enable quantum sensor-specific collaborative learning and analysis while ensuring data privacy? what performance can we expect from such frameworks? and how does the FedQLSTM performance differ from the performance obtained using classical LSTM models in FL frameworks?}

\subsection{Related Works}
The field of QFL was established in \cite{chen2021federated_QML} and \cite{chehimi2021quantum}, in which the first successful implementation of QFL with classical data was presented in \cite{chen2021federated_QML}, and the first implementation of QFL with purely quantum data was presented in \cite{chehimi2021quantum}. A handful of prior works \cite{chehimi2021quantum,li2023pqlm,huang2022quantum,yun2022slimmable,rofougaran2023federated,cao2023linear,chen2022reservoir} attempted to answer the above questions. In particular, the seminal work on QFL with quantum data \cite{chehimi2021quantum} developed the first QFL framework while considering cluster state quantum data that resembles quantum sensing data. However, the model proposed in \cite{chehimi2021quantum} did not consider time-series data such as the one stemming from quantum sensors, and the clients' models in the QFL framework were QCNN models, which performed a binary classification task. Moreover, the work in \cite{li2023pqlm} proposed a multilingual decentralized framework with a portable quantum language model that relies on random VQCs and QLSTM available at a quantum server. The server is fed with temporal text data, and is set to extract and send the word embeddings towards multiple users with classical local machines. Thus, the decentralized framework proposed in \cite{li2023pqlm} did not apply collaborative quantum learning between distant devices, and it considered the complete data to be available at the server. Furthermore, the work in \cite{huang2022quantum} proposed a QFL framework with decentralized data. However, the model proposed in \cite{huang2022quantum} does not consider temporal or time-series data to train its VQCs. Instead, the model in \cite{huang2022quantum} relied on the MNIST dataset for binary classification tasks, as well as a random graph of random qubit samples for an optimization task. Moreover, the authors in \cite{rofougaran2023federated} proposed a QFL framework that leverages the principle of differential privacy to enhance the privacy-preserving characteristics of QFL. While the model in \cite{rofougaran2023federated} has a great potential for improving the security of QSNs, it does not incorporate temporal data nor does it consider a VQC that suits time-series data, like QLSTM.  On another hand, the work in \cite{cao2023linear} considered a QLSTM model, that is enhanced by an additional linear layer before and after the VQCs, for the task of carbon price forecasting. However, the model in \cite{cao2023linear} is a centralized model that does not allow for collaborative training between distributed QLSTM models. Finally, to enhance the training efficiency of QRNN models, the classical principle of reservoir computing was leveraged in \cite{chen2022reservoir}, where the QRNN was trained as a reservoir. However, the QRNN quantum parameters were randomly initialized and fixed in \cite{chen2022reservoir}, and only the final classical layer of the QRNN architecture was trained. Moreover, the model proposed in \cite{chen2022reservoir} is a centralized model that does not incorporate any distributed learning. 

In summary, despite the surge of literature in the field of QFL \cite{chehimi2021quantum,li2023pqlm,huang2022quantum,yun2022slimmable,rofougaran2023federated,cao2023linear,chen2022reservoir}, this prior work does not incorporate QLSTM models in distributed scenarios. Moreover, there are no QFL frameworks that operate on temporal data that can be leveraged in QSN scenarios to perform tasks like function approximation.


\subsection{Contributions}
The main contribution of this paper is a novel QFL framework, dubbed \emph{FedQLSTM}, that enables collaborative learning between multiple QLSTM models in a distributed manner. The FedQLSTM framework has direct applications in distributed QSNs as it operates on temporal data suitable for quantum sensors. After defining the technical foundations of this framework, we study three use cases in which the proposed FedQLSTM framework is leveraged in function approximation tasks based on classical data embedded into quantum states and associated with distributed quantum sensing scenarios and QSNs. We conduct extensive experiments to validate the effectiveness of the proposed FedQLSTM framework. Simulation results show that the proposed FedQLSTM framework achieves a faster convergence compared to FL frameworks with classical LSTM, achieving minimal overall local computations under one local training epoch. In particular, the FedQLSTM framework is shown to achieve a 25-33\% reduction in the number of communication rounds needed until convergence, under only one local training epoch when compared with FL frameworks with classical LSTM models.

The rest of this paper is organized as follows. Section \ref{sec_preliminaries} briefly provides necessary preliminaries on VQCs and QFL, which are necessary to understand the proposed FedQLSTM framework. Next, Section \ref{sec_FedQLSTM} describes the proposed FedQLSTM framework, its constituents, and an algorithmic representation of its operation. Section \ref{sec_usecases} introduces three function approximation use cases of the FedQLSTM framework and provides simulation results to validate its performance. Finally, some conclusions and a future outlook are discussed in Section \ref{sec_conclusion}.




\section{Preliminaries}\label{sec_preliminaries}
\subsection{VQCs}\label{sec_VQCs}
\begin{figure}[t!]
\begin{center}
\includegraphics[width=0.9\textwidth]{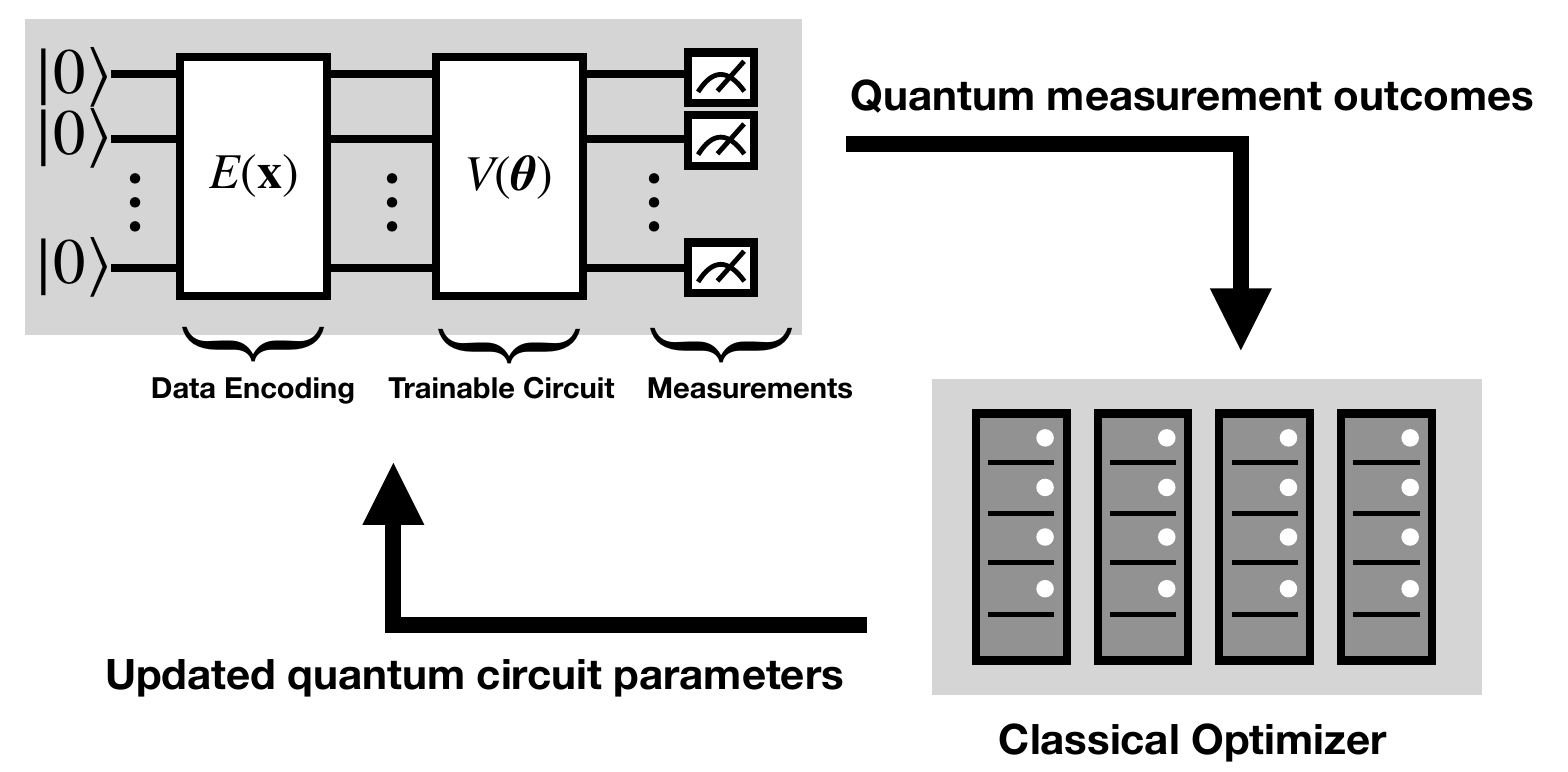}\vspace{-0.1cm}
\caption{Architecture of a hybrid VQC.}
\label{fig_VQC}
\end{center}\vspace{-0.5cm}
\end{figure}
VQCs represent a cutting-edge intersection of quantum mechanics and computational optimization, presenting a quantum analog to classical neural networks. In general, VQCs are hybrid quantum models that incorporate three main elements, as shown in Fig. \ref{fig_VQC}. The first part of a VQC is the data \emph{encoding} phase $E(\boldsymbol{x})$, wherein classical data $\boldsymbol{x}$ is encoded into quantum states. There are various quantum data encoding techniques, each suitable for a different type of quantum algorithms and hardware. Among these, amplitude encoding and variational encoding are the most useful \cite{NIPS2019_quantum_embeddings}. Amplitude encoding maps classical information onto the amplitudes of a quantum state, building on the exponential nature of a quantum system. Under amplitude encoding, an $N$-dimensional input vector requires only $\log_{2}N$ qubits to represent. However, the state preparation circuit depth grows quickly, which is not suitable for existing noisy intermediate-scale quantum (NISQ) devices \cite{mottonen2004transformation}. Variational encoding, on the other hand, uses a sequence of single-qubit parameterized gates whose parameters are adjusted to represent the data within the state of qubits. 

After data embedding, the second element is the \emph{variational} part $V(\boldsymbol{\theta})$ which involves the application of a series of parameterized quantum gates with parameters $\boldsymbol{\theta}$. These gates are unitary transformations that manipulate encoded quantum states, enabling the execution of quantum algorithms. They include the rotation gates $R_x$, $R_y$, and $R_z$ which rotate qubits around the respective axes of the Bloch sphere. The rotation angles are classical learnable parameters, represented by $\boldsymbol{\theta}$, which can be updated via gradient-based or gradient-free classical optimization algorithms \cite{chen2020variational,chen2022variational}. The combination of these gates in sequences forms a learnable quantum circuit, capable of learning the target distribution or tasks. The final element of a VQC is the \emph{measurement} process, which extracts classical information from the quantum system. In particular, the adopted measurement operations in this paper are represented by the Pauli-$Z$ expectation values. The determination of expectation values can be ascertained through analytical methodologies in the context of simulation software, while in the case of real quantum devices, it can be achieved through multiple samplings. The extracted information can be further processed by other VQC or classical operations.
Accordingly, the mathematical expression of a general VQC operation, for a classical input data $\boldsymbol{x}$, is $\overrightarrow{f(\boldsymbol{x} ; \boldsymbol{\theta})}=\left(\left\langle\hat{Z}_1\right\rangle, \cdots,\left\langle\hat{Z}_N\right\rangle\right)$ , where $\left\langle\hat{Z}_{k}\right\rangle =\left\langle 0\left|E^{\dagger}(\boldsymbol{x})V^{\dagger}(\boldsymbol{\theta}) \hat{Z_{k}} V(\boldsymbol{\theta})E(\boldsymbol{x})\right| 0\right\rangle$.
Previous research findings indicate that VQCs exhibit enhanced expressiveness in contrast to classical neural networks~\cite{sim2019expressibility, lanting2014entanglement, du2018expressive, abbas2021power}. In addition, the work in~\cite{caro2022generalization} have demonstrated the efficient training of VQCs with smaller size of datasets.

\subsection{QFL}
FL is a distributed ML framework in which multiple clients train a similar ML model in a collaborative manner with the help of a central server \cite{chen2021distributed}. In FL, the clients do not share the training data with the server but rather send the learning parameters of their local ML models to the server, which then aggregates the learning parameters, updates them, then sends the updated global parameters back to the corresponding clients. This process is repeated over multiple communication rounds until a certain error threshold is achieved. In this regard, QFL corresponds to the adaptation of classical FL setups to include QML models. This adaptation is particularly emphasized in the incorporation of VQC-based quantum neural network models \cite{chehimi2023foundations,ren2023towards}. Since QML models, particularly VQCs, have classical learnable parameters, defined as $\boldsymbol{\theta}$, which can be classically optimized, then a classical communication network can be employed to share such parameters between clients and servers in QFL frameworks \cite{chehimi2021quantum}. Henceforth, we can achieve distributed quantum learning between different quantum devices, e.g., quantum sensors, without the need for developing well-established quantum communication networks \cite{chehimi2022physics,chehimi2023scaling,chehimi2023matching}.

\section{FedQLSTM Framework}\label{sec_FedQLSTM}
\begin{figure}[t!]
\begin{center}
\includegraphics[width=0.9\textwidth]{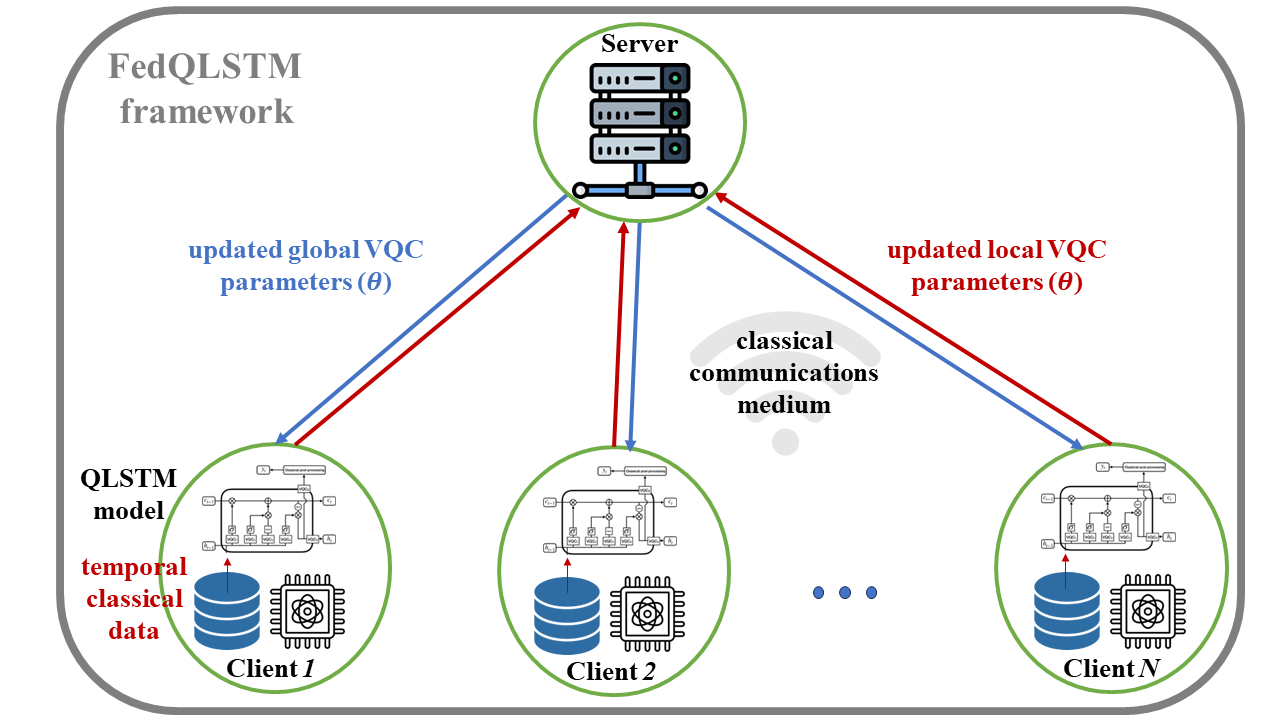}
\caption{Proposed FedQLSTM Framework. }
\label{fig_FedQLSTM}
\end{center}\vspace{-0.3cm}
\end{figure}
\begin{figure}[t!]
\begin{center}
\includegraphics[width=0.65\textwidth]{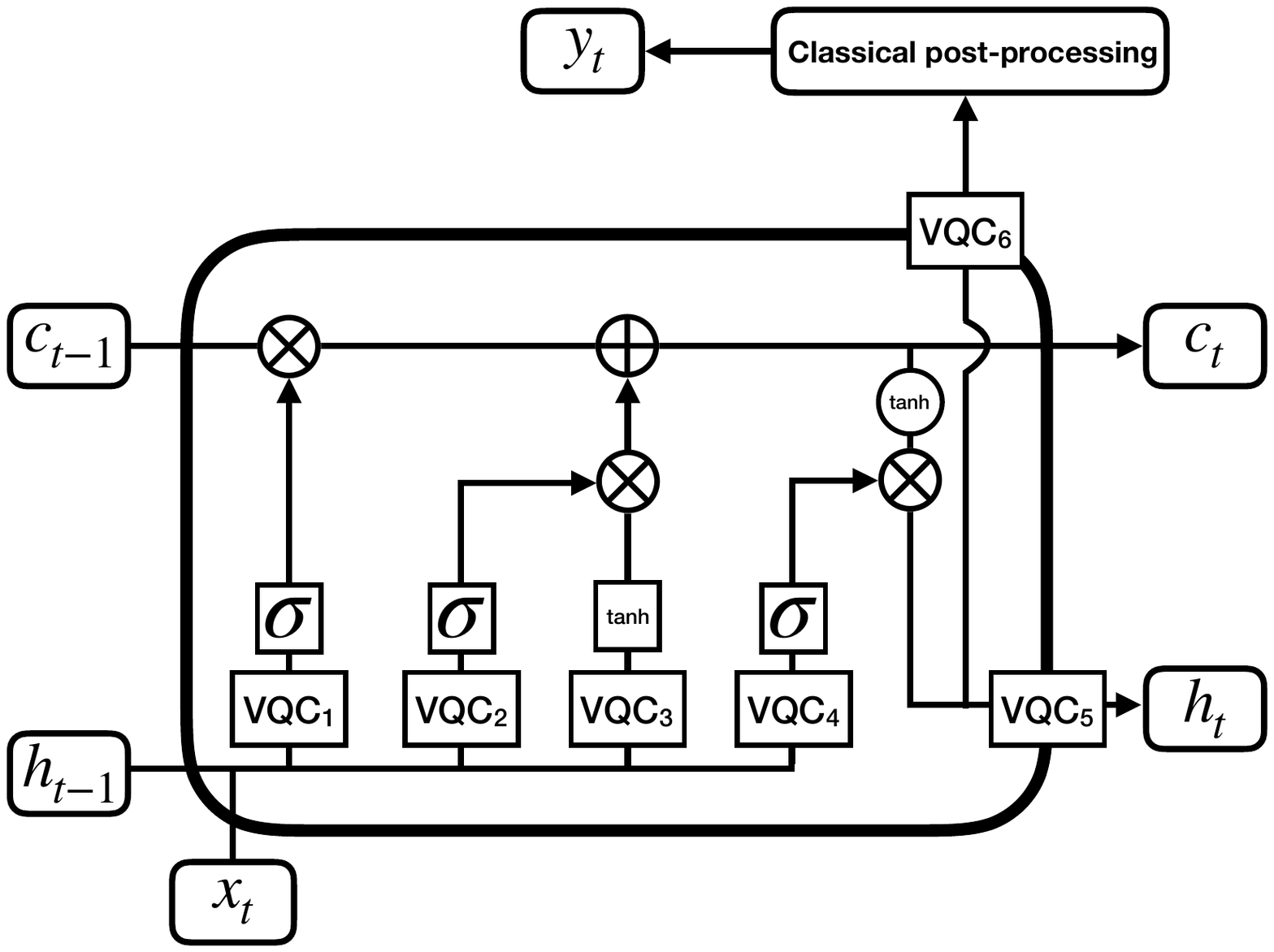}
\caption{Detailed architecture of a hybrid QLSTM model.}
\label{fig_QLSTM}
\end{center}
\end{figure}

The proposed FedQLSTM framework will bring forth a novel integration between federated quantum machine learning frameworks \cite{chen2021federated_QML,chehimi2021quantum} and QLSTM models \cite{chen2022quantum}. In the proposed FedQLSTM framework, shown in Fig.~\ref{fig_FedQLSTM}, we consider a network of $N$ \emph{clients}, with quantum capabilities. Each client is equipped with a local QLSTM model, as shown in Fig.~\ref{fig_QLSTM}. All clients are connected to a \emph{server} through classical communications. Each client has its own local temporal classical data $\boldsymbol{x}$ that is not to be shared with other clients in the network, and is used to locally train the client's QLSTM model. In our simulation, a fixed number of random participating clients will be selected to perform local training and share their trained models during each communication round. The server then aggregates the parameters from the different participating clients and share the aggregated models with all the clients in the network.

\begin{figure}[h]
\begin{center}
\includegraphics[width=0.85\textwidth]{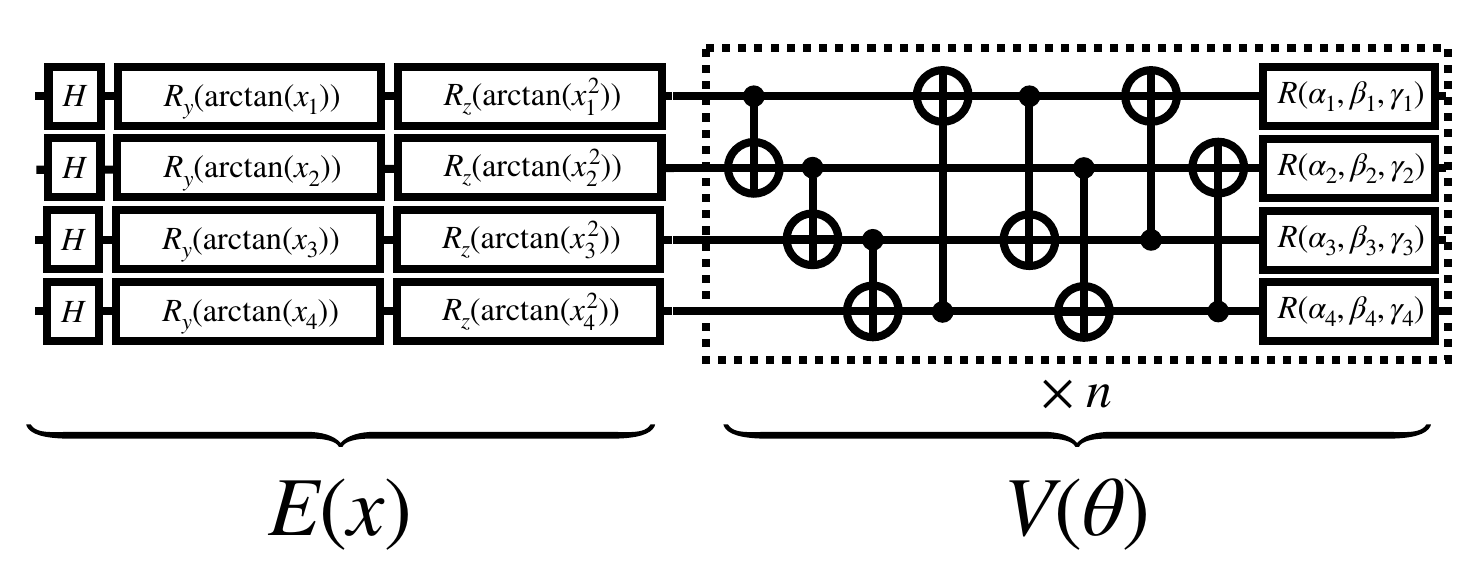}\vspace{-0.1cm}
\caption{VQC used in the QLSTM model.}
\label{fig:vqc_detail}
\end{center}\vspace{-0.4cm}
\end{figure}

There are six VQCs in a QLSTM model. The architecture of these VQCs are shown in Fig.~\ref{fig:vqc_detail}. The encoding circuit $E(\boldsymbol{x})$ includes $R_{y}$ and $R_{z}$ rotations with the rotation angles calculated from the input values. The learnable or trainable circuit (shown as dashed box in Fig.~\ref{fig:vqc_detail}) includes CNOT gates to entangle information from nearby qubits and general unitary gates $R(\alpha, \beta, \gamma)$ with parameters optimized by 
classical gradient-based optimizers. The learnable circuit block can repeat several times to increase the number of parameters. Here, we consider the repeat to be $n=2$. 

The operation of QLSTM \cite{chen2022quantum} shown in Fig.~\ref{fig_QLSTM} can be summarized as follows. At every time step $t$, the input $x_t$ is concatenated with the hidden state $h_{t-1}$ and then the concatenated vector $[x_t, h_{t-1}]$ is sent into the QLSTM cell. Inside the QLSTM cell, the $VQC_{1}$ to $VQC_{4}$ will individually process the vector $[x_t, h_{t-1}]$. This can be done because $[x_t, h_{t-1}]$ is a classical vector so there is no no-cloning issue. Once the VQCs process the input vector $[x_t, h_{t-1}]$, the outputs from each VQC will pass through the corresponding activation functions and then multiple element-wise operations will be carried out. Additional operations of $VQC_{5}$ and $VQC_{6}$ will be carried out to generate the results of $h_{t}$ and $y_{t}$. The output from the QLSTM cell are updated hidden state $h_{t}$, cell state $c_{t}$ and the final output $y_{t}$. Here the hidden state $h_{t}$ and cell state $c_{t}$ will be used again in the next time-step $t+1$, and $y_{t}$ will be collected to calculate the loss. Here, there is a trainable classical post-processing step for the generation of $y_{t}$ to provide a better function fitting. The internal states such as the hidden states $h_{t}$ and cell states $c_{t}$ of every QLSTM are updated based on the input and output from previous time-steps in the sequence, allowing the model to remember long-term dependencies in the input data. These interconnected VQCs as well as the trainable classical neural network can be trained in an end-to-end manner with backpropagation in which the quantum gradients are derived via parameter-shift rule \cite{mitarai2018quantum}.
%
%


The detailed description of the learning process in the proposed FedQLSTM model is described in Algorithm \ref{alg:federated_qlstm}.

 In summary, the proposed FedQLSTM architecture is novel in the sense that it is the first FL framework that incorporates QLSTM models with VQCs. Particularly, this model takes as input classical temporal data that is embedded into quantum states. This framework is useful for several applications like classification and function approximation. Next, we study the application of the proposed FedQLSTM framework in function approximation tasks. In particular, we consider three use cases relevant to quantum sensing, where FedQLSTM can provide an enhanced performance compared to a FL framework with classical LSTM models.

\section{\label{sec_usecases}FedQLSTM Use Cases: Simulation Results and Analysis}
The proposed FedQLSTM framework can be used to perform several tasks, among which function approximation is of great importance. Particularly, function approximation is a crucial step in many machine learning tasks, such as classification, regression, and prediction. By approximating the underlying function that generates the classical input data, the proposed FedQLSTM framework can perform such tasks on decentralized data without compromising the data privacy, while analyzing complex real-world phenomena. In this section, we introduce three important functions that can be approximated in a decentralized manner using the FedQLSTM framework. We then analyze their relevance to quantum sensing technologies, and conduct extensive simulations to compare the performance of the FedQLSTM framework with that of an FL framework with classical LSTM models.

To effectively measure the convergence of our proposed FedQLSTM framework and its classical federated LSTM counterpart, we consider the \emph{sliding-window averaging criterion}, which is based on statistical measures of the loss vector over a specified window of iteration steps \cite{mcmahan2017communication}. We computed the sequential average and standard deviation of this loss vector within each window. Convergence is considered to be achieved at the point where both the average and standard deviation of the sliding window fall below predefined thresholds, indicating stable and minimal variations in the model's performance. If these thresholds were not achieved, the sliding process continues until all iterations are completed, with the total number of iterations being recorded.

\begin{algorithm}[H]
\small
\caption{Federated QLSTM Training}
\label{alg:federated_qlstm}
\begin{algorithmic}[1]
\State \textbf{Input:}\vspace{-0.2cm}
\begin{itemize}\small
\item $K$ federated clients with local datasets ${D_1, D_2, ..., D_K}$
\item $M$ total training iterations (communication rounds)
\item $B$ batch size
\item $E$ number of epochs per training iteration
\item $lr$ learning rate
\item $d$ number of input dimensions
\item $h$ number of hidden dimensions
\item $q$ number of quantum gates
\item $S(\theta)$: QLSTM model with parameters $\theta$
\end{itemize}\vspace{-0.2cm}
\State \textbf{Output:} Global QLSTM model $\theta_{global}$
\State Initialize $\theta_{global}$
\For{iteration $m = 1$ to $M$}
\State Select a subset of clients $S_m \subseteq {1,2,...,K}$
\For{each client $k \in S_m$ in parallel}
\State Initialize $\theta_k = \theta_{global}$
\For{epoch $e = 1$ to $E$}
\State Sample a batch $B_k$ from $D_k$
\State Initialize cell state $c_k$ and hidden state $h_k$ to zero vectors of length $h$
\For{each input sequence $(x_1,...,x_T)$ in $B_k$, and $v_t = [h_{t-1}x_t]$}
\For{timestep $t=1$ to $T$}
\State Compute the input gate $i_t$ and the forget gate $f_t$:
\begin{align*}
f_t &= \sigma(VQC_1(v_t))\\
i_t &= \sigma(VQC_2(v_t))
\end{align*}
\State Compute the candidate cell state $\tilde{c}t$ and update the cell state:
\begin{align*}
\tilde{C}_t &= \tanh(VQC_3(v_t)) \\
c_t &= i_t \circ  \tilde{C}_t + f_t \circ c_{t-1}
\end{align*}
\State Compute the output gate $o_t$ and update the hidden state:
\begin{align*}
o_t &= \sigma(VQC_4(v_t)) \\
h_t &= VQC_5(o_t \circ \tanh(c_t))
\end{align*}
\State Compute the output $y_t$ (when $x_{t}$ is the final timestep $t = T$):
\begin{align*}
y_t &= NN(VQC_6(o_t\circ \tanh(c_t)))
\end{align*}
\EndFor
\EndFor
\State Compute the local gradient $\nabla f_k(\theta_k) = \frac{1}{B} \sum_{(x,y) \in B_k} \nabla L(S(x,\theta_k), y)$
\State Update $\theta_k$ using a gradient-based optimizer (SGD as an example):
\begin{equation*}
\theta_k \gets \theta_k - lr \nabla f_k(\theta_k)\vspace{-0.2cm}
\end{equation*}\vspace{-0.2cm}
\EndFor
\EndFor
\State Aggregate the updated models from each client and update the global model:
\begin{equation*}\small
\theta_{global} \gets \frac{1}{|S_m|} \sum_{k \in S_m} \theta_k \vspace{-0.2cm}
\end{equation*}\vspace{-0.2cm}
\EndFor
\State \textbf{return} $\theta_{global}$
\end{algorithmic}
\end{algorithm}


To compare the performance obtained by the different frameworks, we measure the number of communication rounds needed until convergence, and use it to calculate the \emph{overall local computations} needed until convergence. This measure is defined as  the number of communication rounds needed until convergence multiplied by the number of local training epochs performed before each communication round. Since the clients have limited computing capabilities and the cost of local computations is significantly larger than 
the communication cost, an optimal performance is defined as one that incorporates the minimal amount of overall local computations needed until convergence.

We define the following \emph{default setup} for the simulation parameters. First, we consider the number of participating clients in the federated frameworks to be 5, where each client has a QLSTM model with 4 qubits. Moreover, each user has around 3,000 data samples, split at a ratio of 67\% to 33\% between training and testing data. The adopted optimizer is the RMSprop optimizer, with a learning rate of 0.01 and a batch size of 4. The sliding-window is of width 5, and a 1\% error margin is given for the obtained sliding average. Unless stated otherwise, these parameters are used throughout all simulations.

\subsection{Use Case 1: Bessel Functions}
Bessel functions, specifically of the first kind defined as $J_\alpha(x)$, are solutions to the Bessel differential equation, which is a second-order linear differential equation of the form:
\begin{equation}
    x^{2} \frac{d^{2} y}{d x^{2}} + x \frac{d y}{d x} + (x^{2} - \alpha^{2}) y = 0,
\end{equation}
where ($\alpha$) represents the order of the Bessel function. The solution to this equation is given by the series expansion:
\begin{equation}
    J_{\alpha}(x) = \sum_{m=0}^{\infty} \frac{(-1)^{m}}{m! \, \Gamma(m+\alpha+1)} \left(\frac{x}{2}\right)^{2m+\alpha},
\end{equation}
where ($\Gamma(x)$) is the Gamma function, extending the factorial to real and complex numbers \cite{chen2022quantum}.

Bessel functions play a fundamental role in analyzing the performance of distributed QSNs. In particular, Bessel functions can emerge as part of the quantum Fisher information metric when enhancing the measurement precision of multi-mode optical distributed quantum sensors. For instance, the quadratic dependence on the number of modes in the optical multi-mode quantum sensors can be directly measured through quantum Fisher information that is based on Bessel functions of the first kind \cite{pelayo2023distributed}.

\begin{figure}[!t]
    \centering
    \begin{subfigure}[b]{0.47\textwidth}
        \centering
        \includegraphics[width=\textwidth]{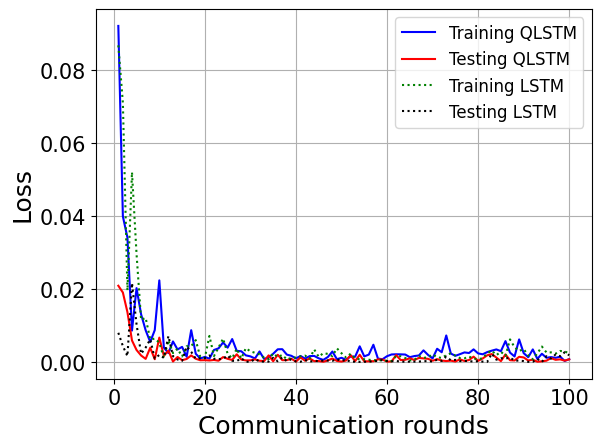}
        \caption{Bessel function approximation losses in the first local training epoch as the number of communication rounds increases.}
        \label{fig_FedQLSTM_bessel_losses}
    \end{subfigure}
    \hspace{0.5cm} 
    \begin{subfigure}[b]{0.47\textwidth}
        \centering
        \includegraphics[width=\textwidth]{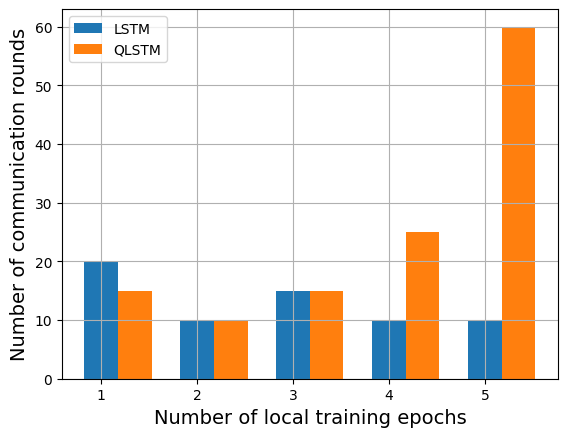}
        \caption{Number of communication rounds needed until convergence for different numbers of local training epochs.}
        \label{fig_FedQLSTM_bessel_convergence}
    \end{subfigure}
    \hspace{0.5cm} 
    \begin{subfigure}[b]{0.5\textwidth}
        \centering
        \includegraphics[width=\textwidth]{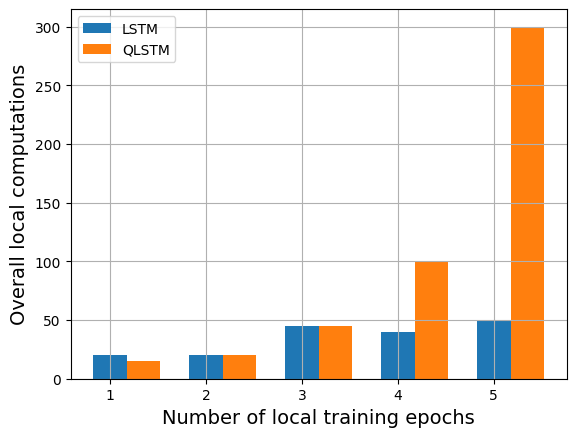}
        \caption{Number of overall local computations needed until convergence for different numbers of local training epochs.}
        \label{fig_FedQLSTM_bessel_overall}
    \end{subfigure}
    \caption{Bessel function approximation performance of FedQLSTM framework vs an FL framework with classical LSTM.}\vspace{-0.5cm}
    \label{fig_bessel_approx}
\end{figure}

\subsubsection{Bessel Function Approximation}
First, we perform the task of approximating a Bessel function of the second order using our proposed FedQLSTM framework and an FL framework with classical LSTM models. In Fig.~\ref{fig_FedQLSTM_bessel_losses}, we show the progress of training and testing errors as the number of communication rounds is increased. From Fig.~\ref{fig_FedQLSTM_bessel_losses}, it is observed that the FedQLSTM framework, despite experiencing some fluctuations in its training, particularly during the initial communication rounds, demonstrates stable performance with respect to testing losses. Furthermore, in Fig.~\ref{fig_FedQLSTM_bessel_convergence}, we show the number of communication rounds needed until convergence for different numbers of local training epochs. We observe from Fig.~\ref{fig_FedQLSTM_bessel_convergence} that, considering only one local training epoch, the FedQLSTM framework requires only 15 communication rounds, thus reducing the number of communication rounds needed till convergence by up to 25\% compared to its classical counterpart. In addition, Fig.~\ref{fig_FedQLSTM_bessel_convergence} shows that both the FedQLSTM and the classical frameworks require fewer number of communication rounds (particularly 10 rounds) until convergence when applying two local training epochs. However, when calculating the overall local computations needed until convergence, as shown in Fig.~\ref{fig_FedQLSTM_bessel_overall}, we observe that the FedQLSTM framework with only one local training epoch requires the minimal amount of overall local computations needed by the clients until convergence. In particular, the FedQLSTM framework with one local training epoch results in saving up to 25\% of the overall local computations needed till convergence compared to the case where two local training epochs are performed. Thus, the FedQLSTM framework with only one local training epoch achieves the optimal convergence that minimizes the overall local computations while being superior to the classical counterpart FL frameworks.

\begin{figure}\vspace{-0.3cm}
\begin{center}
\includegraphics[width=0.6\textwidth]{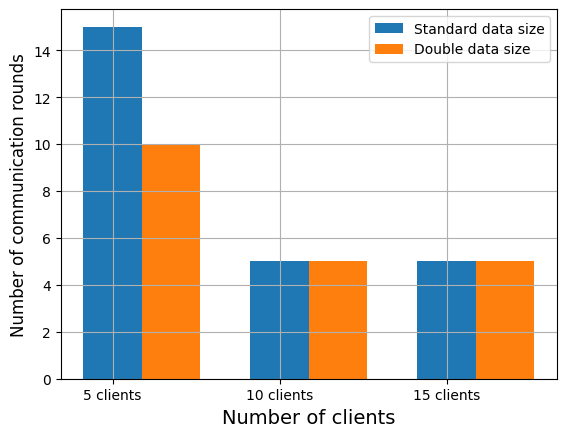}
\caption{Convergence time of the FedQLSTM framework in approximating the Bessel function as the number of clients varies.}
\label{fig_num_users}
\end{center}\vspace{-0.5cm}
\end{figure}

Next, in Fig.~\ref{fig_num_users}, we study the performance of the FedQLSTM framework with one local training epoch in approximating the Bessel function as the number of clients participating in the federated setup increases. From Fig.~\ref{fig_num_users}, we observe that as the number of clients participating in the federated training process increases, the available training data becomes abundant, and thus, the convergence time becomes faster. In particular, when the number of clients increases from 5 to 10 clients, the resulting convergence time is reduced by more than 50\%. Fig.~\ref{fig_num_users} also investigates the effect of doubling the size of each client's dataset in Fig.~\ref{fig_num_users}. In particular, when the number of clients is small, e.g., 5 clients, we observe from Fig.~\ref{fig_num_users} that doubling the size of clients' datasets results in 33\% reduction in the convergence time. Additionally, we observe from Fig.~\ref{fig_num_users} that as the available client data becomes abundant, further increasing the number of users or the data size does not lead to considerable enhancements in the convergence time.  

\subsection{Use Case 2: Delayed Quantum Control}
Next, we examine a quantum feedback system with a delayed control featuring a qubit linked to a semi-infinite waveguide, which terminates at a mirror that fully reflects photons. This setup can be framed as an open quantum system (OQS) in which the waveguide acts as the surrounding environment for the qubit. Notably, this system displays non-Markovian characteristics, particularly when the distance between the qubit and mirror is an integer multiple of the qubit's primary wavelength \cite{fang2018non}. The interaction between the qubit, the reflective surface, and the delayed quantum control of the feedback system lead to the creation of a bound state in the continuum (BIC), trapping some photons in the space between the qubit and the mirror \cite{calajo2019exciting}. By dynamically altering the qubit's frequency, this system allows for the release and subsequent detection of these trapped photons through output field intensity measurements \cite{tufarelli2013dynamics}.

\begin{figure}[!t]
    \centering
    \begin{subfigure}[b]{0.45\textwidth}
        \centering
        \includegraphics[width=\textwidth]{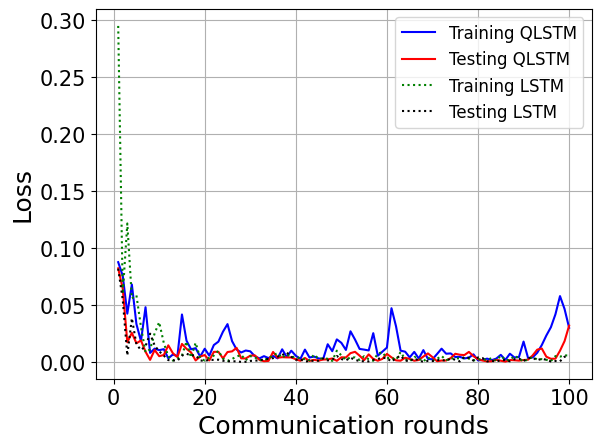}
        \caption{Delayed quantum control sinusoidal function approximation losses in the first local training epoch as the number of communication rounds increases.}
        \label{delayed_first_epoch}
    \end{subfigure}
    \hspace{0.5cm} 
    \begin{subfigure}[b]{0.48\textwidth}
        \centering
        \includegraphics[width=\textwidth]{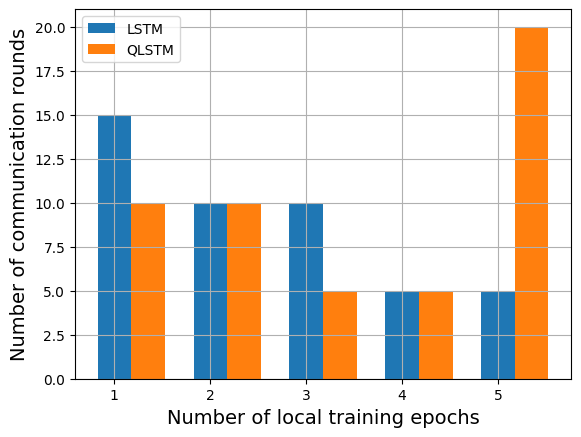}
        \caption{Number of communication rounds needed until convergence for different numbers of local training epochs.}
        \label{delayed_convergence}
    \end{subfigure}
        \hspace{0.5cm} 
    \begin{subfigure}[b]{0.5\textwidth}
        \centering
        \includegraphics[width=\textwidth]{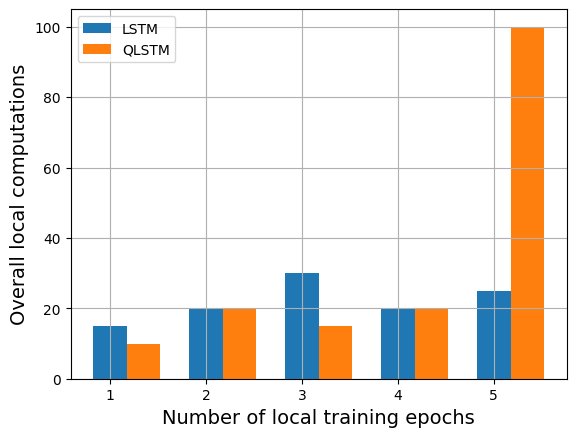}
        \caption{Number of overall local computations needed until convergence for different numbers of local training epochs.}
        \label{fig_FedQLSTM_control_overall}
    \end{subfigure}
    \caption{Delayed quantum control function approximation performance of FedQLSTM framework vs an FL framework with classical LSTM.}
    \label{fig_Delayed_qtm_ctrl}
\end{figure}

\subsubsection{Delayed Quantum Control Function Approximation}
In order to approximate the process of delayed quantum control of a quantum feedback system, we need to model the alterations of the qubit frequency. In this regard, in order for the resulting average qubit frequency after the alteration process to satisfy the resonant condition, we model the variations of the qubit frequency by a sinusoidal modulation \cite{tufarelli2013dynamics}. Accordingly, we analyze the performance of the FedQLSTM framework in approximating the sinusoidal function. From Fig.~\ref{fig_Delayed_qtm_ctrl}, we observe that the FedQLSTM framework with only one local training epoch converges with a minimal amount of overall computations compared to the FL framework with classical LSTM. Moreover, we observe from Fig.~\ref{delayed_convergence} and Fig.~\ref{fig_FedQLSTM_control_overall} that, the FedQLSTM framework manages to save around 33\% of the number of communication rounds and the overall local computations needed until convergence when training for one local epoch, which results in an optimal convergence with minimized overall local computations and major reductions in the number of communication rounds needed until convergence. As such, as was the case when approximating the Bessel function, we observe that the FedQLSTM only requires one local training epoch to optimally approximate the quantum delay sinusoidal function, which further ensures the efficiency of the FedQLSTM framework. This is particularly true since the QLSTM model typically encompasses a significantly smaller number of learning parameters compared to its classical LSTM counterpart \cite{chen2022quantum}.

\subsection{Use Case 3: Struve Function in Quantum Sensors} 
The majority of technologies used for physically implementing quantum states result in qubits that suffer from fragility and short lifetime. An alternative promising approach to overcome such challenges is by developing qubits with mechanical oscillations between quantum mechanical states. Such an approach would enhance the quality of quantum computations and increase the ability to practically manipulate quantum sensors. An example technology is realized through coupling a suspended carbon nanotube to a double quantum dot \cite{pistolesi2021proposal1}. In this regard, an important function to study and model the quantum dynamics of the excitations in carbon nanotubes \cite{pedersen2003variational} and the decoherence of spins \cite{shao1998decoherent} is the \emph{Struve function}. 

Struve functions represent the solution of the following differential equation \cite{olver2010nist}: 
\begin{equation}
    \frac{\text{d}^2w}{\text{d}z^2} + \frac{1}{z} \frac{\text{d}w}{\text{d}z} + (1 - \frac{\nu^2}{z^2})w = \frac{(\frac{1}{2}z)^{\nu - 1}}{\sqrt{\pi}\Gamma(\nu + \frac{1}{2})},
\end{equation}
which is a non-homogeneous Bessel's differential equation. The Struve function is represented by the following power series expansion \cite{olver2010nist}:
\begin{equation}
    H_\nu(z) = (\frac{1}{2}z)^{\nu + 1} \sum_{n=0}^{\infty}{\frac{(-1)^n(\frac{1}{2}z)^{2n}}{\Gamma(n+\frac{3}{2})\Gamma(n+\nu+\frac{3}{2}}},
\end{equation}
where $\nu$ is the order of the Struve function.

Another important use case of the Struve function is for control of the nitrogen-vacancy (NV) spin in diffusing nanodiamonds. Particularly, when performing ensemble-averaged measurements that involve several NV centers in a single crystal, when averaged over all possible Rabi frequencies, incorporate the Struve function as a fundamental element of the mathematical modeling \cite{maclaurin2013nanoscale}. 

\begin{figure}[!t]
    \centering
    \begin{subfigure}[b]{0.45\textwidth}
        \centering
        \includegraphics[width=\textwidth]{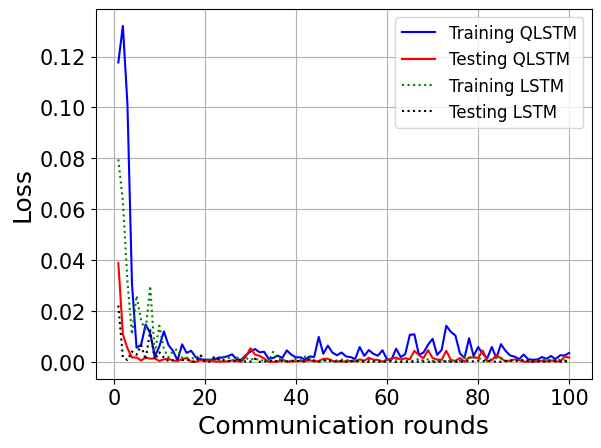}
        \caption{Struve function approximation losses in the first local training epoch as the number of communication rounds increases.}
        \label{struve_first_epoch}
    \end{subfigure}
    \hspace{0.5cm} 
    \begin{subfigure}[b]{0.48\textwidth}
        \centering
        \includegraphics[width=\textwidth]{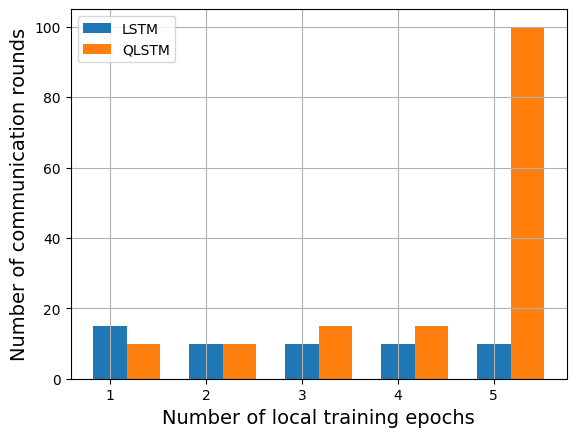}
        \caption{Number of communication rounds needed until convergence for different numbers of local training epochs.}
        \label{struve_convergence}
    \end{subfigure}
            \hspace{0.5cm} 
    \begin{subfigure}[b]{0.5\textwidth}
        \centering
        \includegraphics[width=\textwidth]{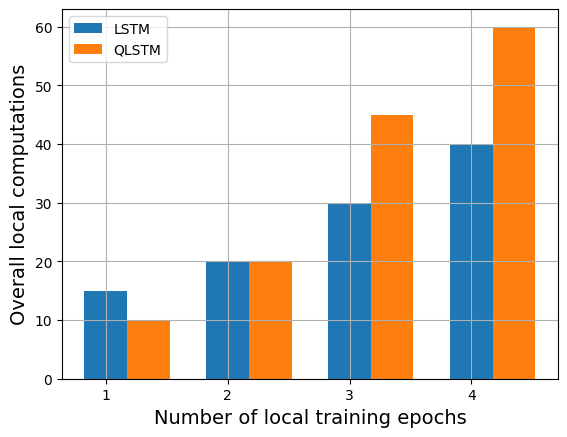}
        \caption{Number of overall local computations needed until convergence for different numbers of local training epochs.}
        \label{fig_FedQLSTM_struve_overall}
    \end{subfigure}
    \caption{Struve function approximation performance of FedQLSTM framework vs an FL framework with classical LSTM.}\vspace{-0.5cm}
    \label{fig_struve_app}
\end{figure}

\subsubsection{Struve function Approximation}
In Fig.~\ref{fig_struve_app}, we show the performance of the FedQLSTM framework against an FL framework with classical LSTM for the task of approximating the Struve function. We observe from Fig.~\ref{fig_struve_app} that the FedQLSTM framework, in one local training epoch, manages to save around 33\% of the number of communication rounds and the overall local computations needed until convergence, compared to the classical counterpart. As such, we can conclude that the training of the FedQLSTM framework with one local epoch achieves an optimal performance that requires minimal overall amount of local computations and, in general, converges faster than FL frameworks with classical LSTM models in terms of the number of communication rounds. This is in compliance with the previous cases of approximating both the Bessel function and the delayed quantum control function.

\begin{figure}[!t]
    \centering
    \includegraphics[width=0.7\textwidth]{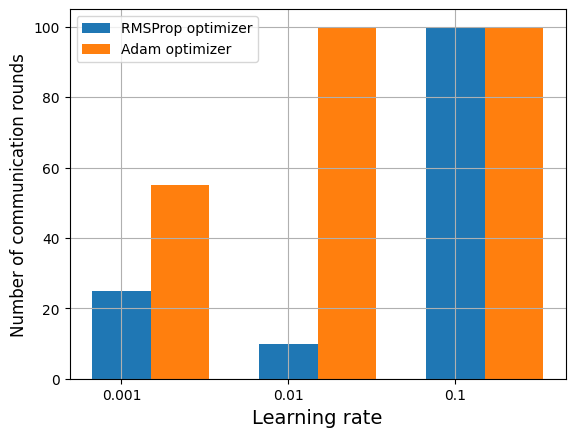}
    \caption{Convergence time in terms of the number of communication rounds for the FedQLSTM framework under different optimizers and learning rates.}
    \label{fig_struve_optimizers}
\end{figure}

Finally, in Fig.~\ref{fig_struve_optimizers}, we compare the performance of the proposed FedQLSTM framework in approximating the Struve function under different optimizers and learning rates when training for one local epoch. From Fig.~\ref{fig_struve_optimizers}, we observe that our selected learning rate of 0.01 and RMSprop optimizer achieve the best Struve function approximation performance. 

\section{Conclusion and Future Outlook}\label{sec_conclusion}

In this paper, we have proposed the first QFL framework with QLSTM models operating on temporal data that resemble the performance of quantum sensors. We have utilized the proposed FedQLSTM framework in performing the task of function approximation, and studied three use cases in this regard. Simulation results confirm that, for most of the use cases, the proposed FedQLSTM framework can achieve a faster convergence rate, saving 25-33\% of the number of communication rounds needed for convergence, under one local training epoch, compared to an FL framework with classical LSTM models. Moreover, under one local training epoch, the FedQLSTM framework results in the minimum number of overall computations needed. Future works must focus on incorporating purely quantum data in the FedQLSTM framework. Such quantum data can be obtained through in-lab experiments and extensive simulations of quantum sensors.

\backmatter



\bibliography{apssamp,bib/tool,bib/struve,bib/delayed_quantum_control,bib/FL,bib/quantum_sensors,bib/qml_examples,bib/QFL,bib/quantum_computing,bib/quantum_communication_networks,bib/vqc}

\providecommand{\noopsort}[1]{}\providecommand{\singleletter}[1]{#1}%

\begin{thebibliography}{50}
\ifx \bisbn   \undefined \def \bisbn  #1{ISBN #1}\fi
\ifx \binits  \undefined \def \binits#1{#1}\fi
\ifx \bauthor  \undefined \def \bauthor#1{#1}\fi
\ifx \batitle  \undefined \def \batitle#1{#1}\fi
\ifx \bjtitle  \undefined \def \bjtitle#1{#1}\fi
\ifx \bvolume  \undefined \def \bvolume#1{\textbf{#1}}\fi
\ifx \byear  \undefined \def \byear#1{#1}\fi
\ifx \bissue  \undefined \def \bissue#1{#1}\fi
\ifx \bfpage  \undefined \def \bfpage#1{#1}\fi
\ifx \blpage  \undefined \def \blpage #1{#1}\fi
\ifx \burl  \undefined \def \burl#1{\textsf{#1}}\fi
\ifx \doiurl  \undefined \def \doiurl#1{\url{https://doi.org/#1}}\fi
\ifx \betal  \undefined \def \betal{\textit{et al.}}\fi
\ifx \binstitute  \undefined \def \binstitute#1{#1}\fi
\ifx \binstitutionaled  \undefined \def \binstitutionaled#1{#1}\fi
\ifx \bctitle  \undefined \def \bctitle#1{#1}\fi
\ifx \beditor  \undefined \def \beditor#1{#1}\fi
\ifx \bpublisher  \undefined \def \bpublisher#1{#1}\fi
\ifx \bbtitle  \undefined \def \bbtitle#1{#1}\fi
\ifx \bedition  \undefined \def \bedition#1{#1}\fi
\ifx \bseriesno  \undefined \def \bseriesno#1{#1}\fi
\ifx \blocation  \undefined \def \blocation#1{#1}\fi
\ifx \bsertitle  \undefined \def \bsertitle#1{#1}\fi
\ifx \bsnm \undefined \def \bsnm#1{#1}\fi
\ifx \bsuffix \undefined \def \bsuffix#1{#1}\fi
\ifx \bparticle \undefined \def \bparticle#1{#1}\fi
\ifx \barticle \undefined \def \barticle#1{#1}\fi
\bibcommenthead
\ifx \bconfdate \undefined \def \bconfdate #1{#1}\fi
\ifx \botherref \undefined \def \botherref #1{#1}\fi
\ifx \url \undefined \def \url#1{\textsf{#1}}\fi
\ifx \bchapter \undefined \def \bchapter#1{#1}\fi
\ifx \bbook \undefined \def \bbook#1{#1}\fi
\ifx \bcomment \undefined \def \bcomment#1{#1}\fi
\ifx \oauthor \undefined \def \oauthor#1{#1}\fi
\ifx \citeauthoryear \undefined \def \citeauthoryear#1{#1}\fi
\ifx \endbibitem  \undefined \def \endbibitem {}\fi
\ifx \bconflocation  \undefined \def \bconflocation#1{#1}\fi
\ifx \arxivurl  \undefined \def \arxivurl#1{\textsf{#1}}\fi
\csname PreBibitemsHook\endcsname

\bibitem[\protect\citeauthoryear{DeBenedictis}{2018}]{debenedictis2018future}
\begin{barticle}
\bauthor{\bsnm{DeBenedictis}, \binits{E.P.}}:
\batitle{A future with quantum machine learning}.
\bjtitle{Computer}
\bvolume{51}(\bissue{2}),
\bfpage{68}--\blpage{71}
(\byear{2018})
\end{barticle}
\endbibitem

\bibitem[\protect\citeauthoryear{Biamonte et~al.}{2017}]{biamonte2017quantum}
\begin{barticle}
\bauthor{\bsnm{Biamonte}, \binits{J.}},
\bauthor{\bsnm{Wittek}, \binits{P.}},
\bauthor{\bsnm{Pancotti}, \binits{N.}},
\bauthor{\bsnm{Rebentrost}, \binits{P.}},
\bauthor{\bsnm{Wiebe}, \binits{N.}},
\bauthor{\bsnm{Lloyd}, \binits{S.}}:
\batitle{Quantum machine learning}.
\bjtitle{Nature}
\bvolume{549}(\bissue{7671}),
\bfpage{195}--\blpage{202}
(\byear{2017})
\end{barticle}
\endbibitem

\bibitem[\protect\citeauthoryear{Benedetti
  et~al.}{2019}]{benedetti2019parameterized}
\begin{barticle}
\bauthor{\bsnm{Benedetti}, \binits{M.}},
\bauthor{\bsnm{Lloyd}, \binits{E.}},
\bauthor{\bsnm{Sack}, \binits{S.}},
\bauthor{\bsnm{Fiorentini}, \binits{M.}}:
\batitle{Parameterized quantum circuits as machine learning models}.
\bjtitle{Quantum Science and Technology}
\bvolume{4}(\bissue{4}),
\bfpage{043001}
(\byear{2019})
\end{barticle}
\endbibitem

\bibitem[\protect\citeauthoryear{Mitarai et~al.}{2018}]{mitarai2018quantum}
\begin{barticle}
\bauthor{\bsnm{Mitarai}, \binits{K.}},
\bauthor{\bsnm{Negoro}, \binits{M.}},
\bauthor{\bsnm{Kitagawa}, \binits{M.}},
\bauthor{\bsnm{Fujii}, \binits{K.}}:
\batitle{Quantum circuit learning}.
\bjtitle{Physical Review A}
\bvolume{98}(\bissue{3}),
\bfpage{032309}
(\byear{2018})
\end{barticle}
\endbibitem

\bibitem[\protect\citeauthoryear{Chen et~al.}{2020}]{chen2020hybrid}
\begin{botherref}
\oauthor{\bsnm{Chen}, \binits{S.Y.-C.}},
\oauthor{\bsnm{Huang}, \binits{C.-M.}},
\oauthor{\bsnm{Hsing}, \binits{C.-W.}},
\oauthor{\bsnm{Kao}, \binits{Y.-J.}}:
Hybrid quantum-classical classifier based on tensor network and variational
  quantum circuit.
arXiv preprint arXiv:2011.14651
(2020)
\end{botherref}
\endbibitem

\bibitem[\protect\citeauthoryear{Schuld et~al.}{2018}]{schuld2018circuit}
\begin{botherref}
\oauthor{\bsnm{Schuld}, \binits{M.}},
\oauthor{\bsnm{Bocharov}, \binits{A.}},
\oauthor{\bsnm{Svore}, \binits{K.}},
\oauthor{\bsnm{Wiebe}, \binits{N.}}:
Circuit-centric quantum classifiers.
arXiv preprint arXiv:1804.00633
(2018)
\end{botherref}
\endbibitem

\bibitem[\protect\citeauthoryear{Chen et~al.}{2022}]{chen2022quantum}
\begin{bchapter}
\bauthor{\bsnm{Chen}, \binits{S.Y.-C.}},
\bauthor{\bsnm{Yoo}, \binits{S.}},
\bauthor{\bsnm{Fang}, \binits{Y.-L.L.}}:
\bctitle{Quantum long short-term memory}.
In: \bbtitle{ICASSP 2022-2022 IEEE International Conference on Acoustics,
  Speech and Signal Processing (ICASSP)},
pp. \bfpage{8622}--\blpage{8626}
(\byear{2022}).
\bcomment{IEEE}
\end{bchapter}
\endbibitem

\bibitem[\protect\citeauthoryear{Bausch}{2020}]{bausch2020recurrent}
\begin{barticle}
\bauthor{\bsnm{Bausch}, \binits{J.}}:
\batitle{Recurrent quantum neural networks}.
\bjtitle{Advances in neural information processing systems}
\bvolume{33},
\bfpage{1368}--\blpage{1379}
(\byear{2020})
\end{barticle}
\endbibitem

\bibitem[\protect\citeauthoryear{Stein et~al.}{2020}]{stein2020qugan}
\begin{botherref}
\oauthor{\bsnm{Stein}, \binits{S.A.}},
\oauthor{\bsnm{Baheri}, \binits{B.}},
\oauthor{\bsnm{Tischio}, \binits{R.M.}},
\oauthor{\bsnm{Mao}, \binits{Y.}},
\oauthor{\bsnm{Guan}, \binits{Q.}},
\oauthor{\bsnm{Li}, \binits{A.}},
\oauthor{\bsnm{Fang}, \binits{B.}},
\oauthor{\bsnm{Xu}, \binits{S.}}:
Qugan: A generative adversarial network through quantum states.
arXiv preprint arXiv:2010.09036
(2020)
\end{botherref}
\endbibitem

\bibitem[\protect\citeauthoryear{Chen et~al.}{2022}]{chen2022variational}
\begin{barticle}
\bauthor{\bsnm{Chen}, \binits{S.Y.-C.}},
\bauthor{\bsnm{Huang}, \binits{C.-M.}},
\bauthor{\bsnm{Hsing}, \binits{C.-W.}},
\bauthor{\bsnm{Goan}, \binits{H.-S.}},
\bauthor{\bsnm{Kao}, \binits{Y.-J.}}:
\batitle{Variational quantum reinforcement learning via evolutionary
  optimization}.
\bjtitle{Machine Learning: Science and Technology}
\bvolume{3}(\bissue{1}),
\bfpage{015025}
(\byear{2022})
\end{barticle}
\endbibitem

\bibitem[\protect\citeauthoryear{Cong et~al.}{2019}]{cong2019quantum}
\begin{barticle}
\bauthor{\bsnm{Cong}, \binits{I.}},
\bauthor{\bsnm{Choi}, \binits{S.}},
\bauthor{\bsnm{Lukin}, \binits{M.D.}}:
\batitle{Quantum convolutional neural networks}.
\bjtitle{Nature Physics}
\bvolume{15}(\bissue{12}),
\bfpage{1273}--\blpage{1278}
(\byear{2019})
\end{barticle}
\endbibitem

\bibitem[\protect\citeauthoryear{Chen et~al.}{2022a}]{chen2020qcnn}
\begin{barticle}
\bauthor{\bsnm{Chen}, \binits{S.Y.-C.}},
\bauthor{\bsnm{Wei}, \binits{T.-C.}},
\bauthor{\bsnm{Zhang}, \binits{C.}},
\bauthor{\bsnm{Yu}, \binits{H.}},
\bauthor{\bsnm{Yoo}, \binits{S.}}:
\batitle{Quantum convolutional neural networks for high energy physics data
  analysis}.
\bjtitle{Physical Review Research}
\bvolume{4}(\bissue{1}),
\bfpage{013231}
(\byear{2022})
\end{barticle}
\endbibitem

\bibitem[\protect\citeauthoryear{Chen et~al.}{2022b}]{chen2022reservoir}
\begin{botherref}
\oauthor{\bsnm{Chen}, \binits{S.Y.-C.}},
\oauthor{\bsnm{Fry}, \binits{D.}},
\oauthor{\bsnm{Deshmukh}, \binits{A.}},
\oauthor{\bsnm{Rastunkov}, \binits{V.}},
\oauthor{\bsnm{Stefanski}, \binits{C.}}:
Reservoir computing via quantum recurrent neural networks.
arXiv preprint arXiv:2211.02612
(2022)
\end{botherref}
\endbibitem

\bibitem[\protect\citeauthoryear{Di~Sipio et~al.}{2022}]{di2022dawn}
\begin{bchapter}
\bauthor{\bsnm{Di~Sipio}, \binits{R.}},
\bauthor{\bsnm{Huang}, \binits{J.-H.}},
\bauthor{\bsnm{Chen}, \binits{S.Y.-C.}},
\bauthor{\bsnm{Mangini}, \binits{S.}},
\bauthor{\bsnm{Worring}, \binits{M.}}:
\bctitle{The dawn of quantum natural language processing}.
In: \bbtitle{ICASSP 2022-2022 IEEE International Conference on Acoustics,
  Speech and Signal Processing (ICASSP)},
pp. \bfpage{8612}--\blpage{8616}
(\byear{2022}).
\bcomment{IEEE}
\end{bchapter}
\endbibitem

\bibitem[\protect\citeauthoryear{Giovannetti
  et~al.}{2004}]{giovannetti2004quantum}
\begin{barticle}
\bauthor{\bsnm{Giovannetti}, \binits{V.}},
\bauthor{\bsnm{Lloyd}, \binits{S.}},
\bauthor{\bsnm{Maccone}, \binits{L.}}:
\batitle{Quantum-enhanced measurements: beating the standard quantum limit}.
\bjtitle{Science}
\bvolume{306}(\bissue{5700}),
\bfpage{1330}--\blpage{1336}
(\byear{2004})
\end{barticle}
\endbibitem

\bibitem[\protect\citeauthoryear{Giovannetti
  et~al.}{2001}]{giovannetti2001quantum}
\begin{barticle}
\bauthor{\bsnm{Giovannetti}, \binits{V.}},
\bauthor{\bsnm{Lloyd}, \binits{S.}},
\bauthor{\bsnm{Maccone}, \binits{L.}}:
\batitle{Quantum-enhanced positioning and clock synchronization}.
\bjtitle{Nature}
\bvolume{412}(\bissue{6845}),
\bfpage{417}--\blpage{419}
(\byear{2001})
\end{barticle}
\endbibitem

\bibitem[\protect\citeauthoryear{Chehimi et~al.}{2023}]{chehimi2023roadmap}
\begin{bchapter}
\bauthor{\bsnm{Chehimi}, \binits{M.}},
\bauthor{\bsnm{Hashash}, \binits{O.}},
\bauthor{\bsnm{Saad}, \binits{W.}}:
\bctitle{The roadmap to a quantum-enabled wireless metaverse: Beyond the
  classical limits}.
In: \bbtitle{2023 Fifth International Conference on Advances in Computational
  Tools for Engineering Applications (ACTEA)},
pp. \bfpage{7}--\blpage{12}
(\byear{2023}).
\bcomment{IEEE}
\end{bchapter}
\endbibitem

\bibitem[\protect\citeauthoryear{Chehimi and
  Saad}{2021}]{chehimi2021entanglement_rate_optimization}
\begin{bchapter}
\bauthor{\bsnm{Chehimi}, \binits{M.}},
\bauthor{\bsnm{Saad}, \binits{W.}}:
\bctitle{Entanglement rate optimization in heterogeneous quantum communication
  networks}.
In: \bbtitle{17th International Symposium on Wireless Communication Systems
  (ISWCS)},
pp. \bfpage{1}--\blpage{6}
(\byear{2021}).
\bcomment{IEEE}
\end{bchapter}
\endbibitem

\bibitem[\protect\citeauthoryear{Kairouz et~al.}{2021}]{kairouz2021advances}
\begin{barticle}
\bauthor{\bsnm{Kairouz}, \binits{P.}},
\bauthor{\bsnm{McMahan}, \binits{H.B.}},
\bauthor{\bsnm{Avent}, \binits{B.}},
\bauthor{\bsnm{Bellet}, \binits{A.}},
\bauthor{\bsnm{Bennis}, \binits{M.}},
\bauthor{\bsnm{Bhagoji}, \binits{A.N.}},
\bauthor{\bsnm{Bonawitz}, \binits{K.}},
\bauthor{\bsnm{Charles}, \binits{Z.}},
\bauthor{\bsnm{Cormode}, \binits{G.}},
\bauthor{\bsnm{Cummings}, \binits{R.}}, \betal:
\batitle{Advances and open problems in federated learning}.
\bjtitle{Foundations and Trends{\textregistered} in Machine Learning}
\bvolume{14}(\bissue{1--2}),
\bfpage{1}--\blpage{210}
(\byear{2021})
\end{barticle}
\endbibitem

\bibitem[\protect\citeauthoryear{Chehimi et~al.}{2023}]{chehimi2023foundations}
\begin{botherref}
\oauthor{\bsnm{Chehimi}, \binits{M.}},
\oauthor{\bsnm{Chen}, \binits{S.Y.-C.}},
\oauthor{\bsnm{Saad}, \binits{W.}},
\oauthor{\bsnm{Towsley}, \binits{D.}},
\oauthor{\bsnm{Debbah}, \binits{M.}}:
Foundations of quantum federated learning over classical and quantum networks.
IEEE Network
(2023)
\end{botherref}
\endbibitem

\bibitem[\protect\citeauthoryear{Chen and Yoo}{2021}]{chen2021federated_QML}
\begin{barticle}
\bauthor{\bsnm{Chen}, \binits{S.Y.-C.}},
\bauthor{\bsnm{Yoo}, \binits{S.}}:
\batitle{Federated quantum machine learning}.
\bjtitle{Entropy}
\bvolume{23}(\bissue{4}),
\bfpage{460}
(\byear{2021})
\end{barticle}
\endbibitem

\bibitem[\protect\citeauthoryear{Chehimi and Saad}{2022}]{chehimi2021quantum}
\begin{bchapter}
\bauthor{\bsnm{Chehimi}, \binits{M.}},
\bauthor{\bsnm{Saad}, \binits{W.}}:
\bctitle{Quantum federated learning with quantum data}.
In: \bbtitle{ICASSP 2022-2022 IEEE International Conference on Acoustics,
  Speech and Signal Processing (ICASSP)},
pp. \bfpage{8617}--\blpage{8621}
(\byear{2022}).
\bcomment{IEEE}
\end{bchapter}
\endbibitem

\bibitem[\protect\citeauthoryear{Li et~al.}{2023}]{li2023pqlm}
\begin{bchapter}
\bauthor{\bsnm{Li}, \binits{S.S.}},
\bauthor{\bsnm{Zhang}, \binits{X.}},
\bauthor{\bsnm{Zhou}, \binits{S.}},
\bauthor{\bsnm{Shu}, \binits{H.}},
\bauthor{\bsnm{Liang}, \binits{R.}},
\bauthor{\bsnm{Liu}, \binits{H.}},
\bauthor{\bsnm{Garcia}, \binits{L.P.}}:
\bctitle{Pqlm-multilingual decentralized portable quantum language model}.
In: \bbtitle{ICASSP 2023-2023 IEEE International Conference on Acoustics,
  Speech and Signal Processing (ICASSP)},
pp. \bfpage{1}--\blpage{5}
(\byear{2023}).
\bcomment{IEEE}
\end{bchapter}
\endbibitem

\bibitem[\protect\citeauthoryear{Huang et~al.}{2022}]{huang2022quantum}
\begin{barticle}
\bauthor{\bsnm{Huang}, \binits{R.}},
\bauthor{\bsnm{Tan}, \binits{X.}},
\bauthor{\bsnm{Xu}, \binits{Q.}}:
\batitle{Quantum federated learning with decentralized data}.
\bjtitle{IEEE Journal of Selected Topics in Quantum Electronics}
\bvolume{28}(\bissue{4}),
\bfpage{1}--\blpage{10}
(\byear{2022})
\end{barticle}
\endbibitem

\bibitem[\protect\citeauthoryear{Yun et~al.}{2022}]{yun2022slimmable}
\begin{botherref}
\oauthor{\bsnm{Yun}, \binits{W.J.}},
\oauthor{\bsnm{Kim}, \binits{J.P.}},
\oauthor{\bsnm{Jung}, \binits{S.}},
\oauthor{\bsnm{Park}, \binits{J.}},
\oauthor{\bsnm{Bennis}, \binits{M.}},
\oauthor{\bsnm{Kim}, \binits{J.}}:
Slimmable quantum federated learning.
arXiv preprint arXiv:2207.10221
(2022)
\end{botherref}
\endbibitem

\bibitem[\protect\citeauthoryear{Rofougaran
  et~al.}{2023}]{rofougaran2023federated}
\begin{botherref}
\oauthor{\bsnm{Rofougaran}, \binits{R.}},
\oauthor{\bsnm{Yoo}, \binits{S.}},
\oauthor{\bsnm{Tseng}, \binits{H.-H.}},
\oauthor{\bsnm{Chen}, \binits{S.Y.-C.}}:
Federated quantum machine learning with differential privacy.
arXiv preprint arXiv:2310.06973
(2023)
\end{botherref}
\endbibitem

\bibitem[\protect\citeauthoryear{Cao et~al.}{2023}]{cao2023linear}
\begin{barticle}
\bauthor{\bsnm{Cao}, \binits{Y.}},
\bauthor{\bsnm{Zhou}, \binits{X.}},
\bauthor{\bsnm{Fei}, \binits{X.}},
\bauthor{\bsnm{Zhao}, \binits{H.}},
\bauthor{\bsnm{Liu}, \binits{W.}},
\bauthor{\bsnm{Zhao}, \binits{J.}}:
\batitle{Linear-layer-enhanced quantum long short-term memory for carbon price
  forecasting}.
\bjtitle{Quantum Machine Intelligence}
\bvolume{5}(\bissue{2}),
\bfpage{1}--\blpage{12}
(\byear{2023})
\end{barticle}
\endbibitem

\bibitem[\protect\citeauthoryear{Garg
  et~al.}{2019}]{NIPS2019_quantum_embeddings}
\begin{bchapter}
\bauthor{\bsnm{Garg}, \binits{D.}},
\bauthor{\bsnm{Ikbal}, \binits{S.}},
\bauthor{\bsnm{Srivastava}, \binits{S.K.}},
\bauthor{\bsnm{Vishwakarma}, \binits{H.}},
\bauthor{\bsnm{Karanam}, \binits{H.}},
\bauthor{\bsnm{Subramaniam}, \binits{L.V.}}:
\bctitle{Quantum embedding of knowledge for reasoning}.
In: \beditor{\bsnm{Wallach}, \binits{H.}},
\beditor{\bsnm{Larochelle}, \binits{H.}},
\beditor{\bsnm{Beygelzimer}, \binits{A.}},
\beditor{\bsnm{Alch\'{e}-Buc}, \binits{F.}},
\beditor{\bsnm{Fox}, \binits{E.}},
\beditor{\bsnm{Garnett}, \binits{R.}} (eds.)
\bbtitle{Advances in Neural Information Processing Systems 32},
pp. \bfpage{5594}--\blpage{5604}
(\byear{2019})
\end{bchapter}
\endbibitem

\bibitem[\protect\citeauthoryear{Mottonen
  et~al.}{2004}]{mottonen2004transformation}
\begin{botherref}
\oauthor{\bsnm{Mottonen}, \binits{M.}},
\oauthor{\bsnm{Vartiainen}, \binits{J.J.}},
\oauthor{\bsnm{Bergholm}, \binits{V.}},
\oauthor{\bsnm{Salomaa}, \binits{M.M.}}:
Transformation of quantum states using uniformly controlled rotations.
arXiv preprint quant-ph/0407010
(2004)
\end{botherref}
\endbibitem

\bibitem[\protect\citeauthoryear{Chen et~al.}{2020}]{chen2020variational}
\begin{barticle}
\bauthor{\bsnm{Chen}, \binits{S.Y.-C.}},
\bauthor{\bsnm{Yang}, \binits{C.-H.H.}},
\bauthor{\bsnm{Qi}, \binits{J.}},
\bauthor{\bsnm{Chen}, \binits{P.-Y.}},
\bauthor{\bsnm{Ma}, \binits{X.}},
\bauthor{\bsnm{Goan}, \binits{H.-S.}}:
\batitle{Variational quantum circuits for deep reinforcement learning}.
\bjtitle{IEEE Access}
\bvolume{8},
\bfpage{141007}--\blpage{141024}
(\byear{2020})
\end{barticle}
\endbibitem

\bibitem[\protect\citeauthoryear{Sim et~al.}{2019}]{sim2019expressibility}
\begin{barticle}
\bauthor{\bsnm{Sim}, \binits{S.}},
\bauthor{\bsnm{Johnson}, \binits{P.D.}},
\bauthor{\bsnm{Aspuru-Guzik}, \binits{A.}}:
\batitle{Expressibility and entangling capability of parameterized quantum
  circuits for hybrid quantum-classical algorithms}.
\bjtitle{Advanced Quantum Technologies}
\bvolume{2}(\bissue{12}),
\bfpage{1900070}
(\byear{2019})
\end{barticle}
\endbibitem

\bibitem[\protect\citeauthoryear{Lanting
  et~al.}{2014}]{lanting2014entanglement}
\begin{barticle}
\bauthor{\bsnm{Lanting}, \binits{T.}},
\bauthor{\bsnm{Przybysz}, \binits{A.J.}},
\bauthor{\bsnm{Smirnov}, \binits{A.Y.}},
\bauthor{\bsnm{Spedalieri}, \binits{F.M.}},
\bauthor{\bsnm{Amin}, \binits{M.H.}},
\bauthor{\bsnm{Berkley}, \binits{A.J.}},
\bauthor{\bsnm{Harris}, \binits{R.}},
\bauthor{\bsnm{Altomare}, \binits{F.}},
\bauthor{\bsnm{Boixo}, \binits{S.}},
\bauthor{\bsnm{Bunyk}, \binits{P.}}, \betal:
\batitle{Entanglement in a quantum annealing processor}.
\bjtitle{Physical Review X}
\bvolume{4}(\bissue{2}),
\bfpage{021041}
(\byear{2014})
\end{barticle}
\endbibitem

\bibitem[\protect\citeauthoryear{Du et~al.}{2018}]{du2018expressive}
\begin{botherref}
\oauthor{\bsnm{Du}, \binits{Y.}},
\oauthor{\bsnm{Hsieh}, \binits{M.-H.}},
\oauthor{\bsnm{Liu}, \binits{T.}},
\oauthor{\bsnm{Tao}, \binits{D.}}:
The expressive power of parameterized quantum circuits.
arXiv preprint arXiv:1810.11922
(2018)
\end{botherref}
\endbibitem

\bibitem[\protect\citeauthoryear{Abbas et~al.}{2021}]{abbas2021power}
\begin{barticle}
\bauthor{\bsnm{Abbas}, \binits{A.}},
\bauthor{\bsnm{Sutter}, \binits{D.}},
\bauthor{\bsnm{Zoufal}, \binits{C.}},
\bauthor{\bsnm{Lucchi}, \binits{A.}},
\bauthor{\bsnm{Figalli}, \binits{A.}},
\bauthor{\bsnm{Woerner}, \binits{S.}}:
\batitle{The power of quantum neural networks}.
\bjtitle{Nature Computational Science}
\bvolume{1}(\bissue{6}),
\bfpage{403}--\blpage{409}
(\byear{2021})
\end{barticle}
\endbibitem

\bibitem[\protect\citeauthoryear{Caro et~al.}{2022}]{caro2022generalization}
\begin{barticle}
\bauthor{\bsnm{Caro}, \binits{M.C.}},
\bauthor{\bsnm{Huang}, \binits{H.-Y.}},
\bauthor{\bsnm{Cerezo}, \binits{M.}},
\bauthor{\bsnm{Sharma}, \binits{K.}},
\bauthor{\bsnm{Sornborger}, \binits{A.}},
\bauthor{\bsnm{Cincio}, \binits{L.}},
\bauthor{\bsnm{Coles}, \binits{P.J.}}:
\batitle{Generalization in quantum machine learning from few training data}.
\bjtitle{Nature communications}
\bvolume{13}(\bissue{1}),
\bfpage{1}--\blpage{11}
(\byear{2022})
\end{barticle}
\endbibitem

\bibitem[\protect\citeauthoryear{Chen et~al.}{2021}]{chen2021distributed}
\begin{barticle}
\bauthor{\bsnm{Chen}, \binits{M.}},
\bauthor{\bsnm{G{\"u}nd{\"u}z}, \binits{D.}},
\bauthor{\bsnm{Huang}, \binits{K.}},
\bauthor{\bsnm{Saad}, \binits{W.}},
\bauthor{\bsnm{Bennis}, \binits{M.}},
\bauthor{\bsnm{Feljan}, \binits{A.V.}},
\bauthor{\bsnm{Poor}, \binits{H.V.}}:
\batitle{Distributed learning in wireless networks: Recent progress and future
  challenges}.
\bjtitle{IEEE Journal on Selected Areas in Communications}
\bvolume{39}(\bissue{12}),
\bfpage{3579}--\blpage{3605}
(\byear{2021})
\end{barticle}
\endbibitem

\bibitem[\protect\citeauthoryear{Ren et~al.}{2023}]{ren2023towards}
\begin{botherref}
\oauthor{\bsnm{Ren}, \binits{C.}},
\oauthor{\bsnm{Yu}, \binits{H.}},
\oauthor{\bsnm{Yan}, \binits{R.}},
\oauthor{\bsnm{Xu}, \binits{M.}},
\oauthor{\bsnm{Shen}, \binits{Y.}},
\oauthor{\bsnm{Zhu}, \binits{H.}},
\oauthor{\bsnm{Niyato}, \binits{D.}},
\oauthor{\bsnm{Dong}, \binits{Z.Y.}},
\oauthor{\bsnm{Kwek}, \binits{L.C.}}:
Towards quantum federated learning.
arXiv preprint arXiv:2306.09912
(2023)
\end{botherref}
\endbibitem

\bibitem[\protect\citeauthoryear{Chehimi and Saad}{2022}]{chehimi2022physics}
\begin{botherref}
\oauthor{\bsnm{Chehimi}, \binits{M.}},
\oauthor{\bsnm{Saad}, \binits{W.}}:
Physics-informed quantum communication networks: A vision towards the quantum
  internet.
IEEE network,
134--142
(2022)
\end{botherref}
\endbibitem

\bibitem[\protect\citeauthoryear{Chehimi et~al.}{2023a}]{chehimi2023scaling}
\begin{bchapter}
\bauthor{\bsnm{Chehimi}, \binits{M.}},
\bauthor{\bsnm{Pouryousef}, \binits{S.}},
\bauthor{\bsnm{Panigrahy}, \binits{N.}},
\bauthor{\bsnm{Towsley}, \binits{D.}},
\bauthor{\bsnm{Saad}, \binits{W.}}:
\bctitle{Scaling limits of quantum repeater networks}.
In: \bbtitle{Proc. of IEEE International Conference on Quantum Computing and
  Engineering (QCE)},
\bconflocation{Bellevue, WA USA}
(\byear{2023})
\end{bchapter}
\endbibitem

\bibitem[\protect\citeauthoryear{Chehimi et~al.}{2023b}]{chehimi2023matching}
\begin{bchapter}
\bauthor{\bsnm{Chehimi}, \binits{M.}},
\bauthor{\bsnm{Simon}, \binits{B.}},
\bauthor{\bsnm{Saad}, \binits{W.}},
\bauthor{\bsnm{Klein}, \binits{A.}},
\bauthor{\bsnm{Towsley}, \binits{D.}},
\bauthor{\bsnm{Debbah}, \binits{M.}}:
\bctitle{Matching game for optimized association in quantum communication
  networks}.
In: \bbtitle{Proc. of IEEE Global Communications Conference (Globecom)},
\bconflocation{Kuala Lumpur, Malaysia}
(\byear{2023})
\end{bchapter}
\endbibitem

\bibitem[\protect\citeauthoryear{McMahan
  et~al.}{2017}]{mcmahan2017communication}
\begin{bchapter}
\bauthor{\bsnm{McMahan}, \binits{B.}},
\bauthor{\bsnm{Moore}, \binits{E.}},
\bauthor{\bsnm{Ramage}, \binits{D.}},
\bauthor{\bsnm{Hampson}, \binits{S.}},
\bauthor{\bsnm{Arcas}, \binits{B.A.}}:
\bctitle{Communication-efficient learning of deep networks from decentralized
  data}.
In: \bbtitle{Artificial Intelligence and Statistics},
pp. \bfpage{1273}--\blpage{1282}
(\byear{2017}).
\bcomment{PMLR}
\end{bchapter}
\endbibitem

\bibitem[\protect\citeauthoryear{Pelayo et~al.}{2023}]{pelayo2023distributed}
\begin{barticle}
\bauthor{\bsnm{Pelayo}, \binits{J.C.}},
\bauthor{\bsnm{Gietka}, \binits{K.}},
\bauthor{\bsnm{Busch}, \binits{T.}}:
\batitle{Distributed quantum sensing with optical lattices}.
\bjtitle{Physical Review A}
\bvolume{107}(\bissue{3}),
\bfpage{033318}
(\byear{2023})
\end{barticle}
\endbibitem

\bibitem[\protect\citeauthoryear{Fang et~al.}{2018}]{fang2018non}
\begin{barticle}
\bauthor{\bsnm{Fang}, \binits{Y.-L.L.}},
\bauthor{\bsnm{Ciccarello}, \binits{F.}},
\bauthor{\bsnm{Baranger}, \binits{H.U.}}:
\batitle{Non-markovian dynamics of a qubit due to single-photon scattering in a
  waveguide}.
\bjtitle{New Journal of Physics}
\bvolume{20}(\bissue{4}),
\bfpage{043035}
(\byear{2018})
\end{barticle}
\endbibitem

\bibitem[\protect\citeauthoryear{Calaj{\'o} et~al.}{2019}]{calajo2019exciting}
\begin{barticle}
\bauthor{\bsnm{Calaj{\'o}}, \binits{G.}},
\bauthor{\bsnm{Fang}, \binits{Y.-L.L.}},
\bauthor{\bsnm{Baranger}, \binits{H.U.}},
\bauthor{\bsnm{Ciccarello}, \binits{F.}}, \betal:
\batitle{Exciting a bound state in the continuum through multiphoton scattering
  plus delayed quantum feedback}.
\bjtitle{Physical review letters}
\bvolume{122}(\bissue{7}),
\bfpage{073601}
(\byear{2019})
\end{barticle}
\endbibitem

\bibitem[\protect\citeauthoryear{Tufarelli
  et~al.}{2013}]{tufarelli2013dynamics}
\begin{barticle}
\bauthor{\bsnm{Tufarelli}, \binits{T.}},
\bauthor{\bsnm{Ciccarello}, \binits{F.}},
\bauthor{\bsnm{Kim}, \binits{M.}}:
\batitle{Dynamics of spontaneous emission in a single-end photonic waveguide}.
\bjtitle{Physical Review A}
\bvolume{87}(\bissue{1}),
\bfpage{013820}
(\byear{2013})
\end{barticle}
\endbibitem

\bibitem[\protect\citeauthoryear{Pistolesi
  et~al.}{2021}]{pistolesi2021proposal1}
\begin{barticle}
\bauthor{\bsnm{Pistolesi}, \binits{F.}},
\bauthor{\bsnm{Cleland}, \binits{A.}},
\bauthor{\bsnm{Bachtold}, \binits{A.}}:
\batitle{Proposal for a nanomechanical qubit}.
\bjtitle{Physical Review X}
\bvolume{11}(\bissue{3}),
\bfpage{031027}
(\byear{2021})
\end{barticle}
\endbibitem

\bibitem[\protect\citeauthoryear{Pedersen}{2003}]{pedersen2003variational}
\begin{barticle}
\bauthor{\bsnm{Pedersen}, \binits{T.G.}}:
\batitle{Variational approach to excitons in carbon nanotubes}.
\bjtitle{Physical Review B}
\bvolume{67}(\bissue{7}),
\bfpage{073401}
(\byear{2003})
\end{barticle}
\endbibitem

\bibitem[\protect\citeauthoryear{Shao and
  H{\"a}nggi}{1998}]{shao1998decoherent}
\begin{barticle}
\bauthor{\bsnm{Shao}, \binits{J.}},
\bauthor{\bsnm{H{\"a}nggi}, \binits{P.}}:
\batitle{Decoherent dynamics of a two-level system coupled to a sea of spins}.
\bjtitle{Physical review letters}
\bvolume{81}(\bissue{26}),
\bfpage{5710}
(\byear{1998})
\end{barticle}
\endbibitem

\bibitem[\protect\citeauthoryear{Olver et~al.}{2010}]{olver2010nist}
\begin{bbook}
\bauthor{\bsnm{Olver}, \binits{F.W.}},
\bauthor{\bsnm{Lozier}, \binits{D.W.}},
\bauthor{\bsnm{Boisvert}, \binits{R.F.}},
\bauthor{\bsnm{Clark}, \binits{C.W.}}:
\bbtitle{NIST Handbook of Mathematical Functions Hardback and CD-ROM}.
\bpublisher{Cambridge university press}, \blocation{???}
(\byear{2010})
\end{bbook}
\endbibitem

\bibitem[\protect\citeauthoryear{Maclaurin
  et~al.}{2013}]{maclaurin2013nanoscale}
\begin{barticle}
\bauthor{\bsnm{Maclaurin}, \binits{D.}},
\bauthor{\bsnm{Hall}, \binits{L.}},
\bauthor{\bsnm{Martin}, \binits{A.}},
\bauthor{\bsnm{Hollenberg}, \binits{L.}}:
\batitle{Nanoscale magnetometry through quantum control of nitrogen--vacancy
  centres in rotationally diffusing nanodiamonds}.
\bjtitle{New Journal of Physics}
\bvolume{15}(\bissue{1}),
\bfpage{013041}
(\byear{2013})
\end{barticle}
\endbibitem

\end{thebibliography}

\end{document}